\def\paperID{7768}
\def\confName{CVPR}

\def\authorBlock{
    Pengyuan Wang \footnotemark[1]  \qquad
    Takuya Ikeda \footnotemark[2] \qquad
    Robert Lee \footnotemark[2]  \qquad
    Koichi Nishiwaki \footnotemark[2] \\
    Technical University of Munich \footnotemark[1] \qquad
    Woven by Toyota \footnotemark[2] \\
     {\tt\small pengyuan.wang@tum.de} \qquad
    {\tt\small \{takuya.ikeda, robert.lee, koichi.nishiwaki \}@woven-planet.global}
}

\newif\ifreview 
\newif\ifarxiv \newcommand{\arxiv}{\arxivtrue}
\newif\ifcamera 
\newif\ifrebuttal 

\arxiv %

\pdfoutput=1
\documentclass[10pt,twocolumn,letterpaper]{article}
\usepackage[fleqn]{amsmath}
\ifreview \usepackage[review]{cvpr} \fi
\ifarxiv \usepackage[pagenumbers]{cvpr} \fi
\ifrebuttal \usepackage[rebuttal]{cvpr} \fi
\ifcamera \usepackage{cvpr} \fi

\usepackage{graphicx}	
\usepackage{amsmath}	
\usepackage{amssymb}	
\usepackage{booktabs}
\usepackage{times}
\usepackage{microtype}
\usepackage{epsfig}
\usepackage[table,xcdraw,dvipsnames]{xcolor}
\usepackage{caption}
\usepackage{float}
\usepackage{placeins}
\usepackage{color, colortbl}
\usepackage{stfloats}
\usepackage{enumitem}
\usepackage{tabularx}
\usepackage{xstring}
\usepackage{multirow}
\usepackage{xspace}
\usepackage{url}
\usepackage{subcaption}
\usepackage{xcolor}
\usepackage{pdfpages}
\usepackage[dvipsnames]{xcolor}
\usepackage{MnSymbol,bbding,pifont}
\usepackage[hang,flushmargin]{footmisc}

\ifcamera \usepackage[accsupp]{axessibility} \fi

\ifarxiv  \fi

\newcommand{\R}[1]{{%
    \textbf{%
        \ifstrequal{#1}{1}{\textcolor{red}{R#1}}{%
        \ifstrequal{#1}{2}{\textcolor{blue}{R#1}}{%
        \ifstrequal{#1}{3}{\textcolor{magenta}{R#1}}{%
        \ifstrequal{#1}{4}{\textcolor{teal}{R#1}}{%
                           \textcolor{cyan}{R#1}%
        }}}}%
    }%
}}

\usepackage{xr-hyper}
\usepackage{svg}
\usepackage{pdfpages}
\usepackage{graphicx} 
\usepackage{amsmath}
\usepackage{amssymb}
\usepackage{booktabs}
\usepackage{makecell}
\usepackage{color}
\usepackage{textcomp,gensymb}
\usepackage{sections/mytables}
\usepackage{multirow}

\makeatletter
\newcommand*{\addFileDependency}[1]{
  \typeout{(#1)}
  \@addtofilelist{#1}
  \IfFileExists{#1}{}{\typeout{No file #1.}}
}

\makeatother

\definecolor{cvprblue}{rgb}{0.21,0.49,0.74}
\usepackage[pagebackref,breaklinks,colorlinks,citecolor=cvprblue]{hyperref}
\usepackage[capitalize]{cleveref}
\crefname{section}{Sec.}{Secs.}
\crefname{table}{Table}{Tables}
\crefname{figure}{Fig.}{Figs.}

\begin{document}
\title{GS-Pose: Category-Level Object Pose Estimation via Geometric and Semantic Correspondence}
\author{\authorBlock}
\maketitle

\begin{abstract}
    Category-level pose estimation is a challenging task with many potential applications in computer vision and robotics. Recently, deep-learning-based approaches have made great progress, but are typically hindered by the need for large datasets of either pose-labelled real images or carefully tuned photorealistic simulators. 
    This can be avoided by using only geometry inputs such as depth images to reduce the domain-gap 
    but these approaches suffer from a lack of semantic information, which can be vital in the pose estimation problem. 
    To resolve this conflict, we propose to utilize both geometric and semantic features obtained from a pre-trained foundation model.
    Our approach projects 2D features from this foundation model into 3D for a single object model per category, and then performs matching against this for new single view observations of unseen object instances with a trained matching network. This requires significantly less data to train than prior methods since the semantic features are robust to object texture and appearance. 
    We demonstrate this with a rich evaluation, showing improved performance over prior methods with a fraction of the data required. %
\end{abstract}
\section{Introduction}
Object pose estimation is a fundamental problem in the computer vision and robotics fields.
With the advancement of deep learning methods, various learning-based pose estimation approaches have proven effective for instance-level pose estimation \cite{Peng2019-wx, He2022-wy, He2023-tw, Liu2022-jp, pan2023learning, Nguyen2023-lb, Zhao2022-ej, Wang2020-an, Wang2021-oo, Wang_2021_self6dpp, Iwase2021-fk, Xu2022-ts, Chen2023-cu, Hodan2020-wy, Labbe2020-xp, Ikeda2022-qs}.
Furthermore, recent approaches have extended the pose estimation problem from instance-level to category-level, estimating the pose of unseen object instances within a given category. %

RGB input can be useful for instance-level object pose estimation
since the texture information is highly correlated to the pose and can help resolve ambiguity issues such as axis-symmetry. 
However, for category-level object pose estimation, 
using RGB input can make training data more complicated.
For example, most RGB based category-level approaches require a significant amount of real data with pose annotation or photorealistic synthetic data to cover the large variety of appearances that instances within a single category may have.

\begin{figure}
\centering
\includegraphics[width=0.95\linewidth]{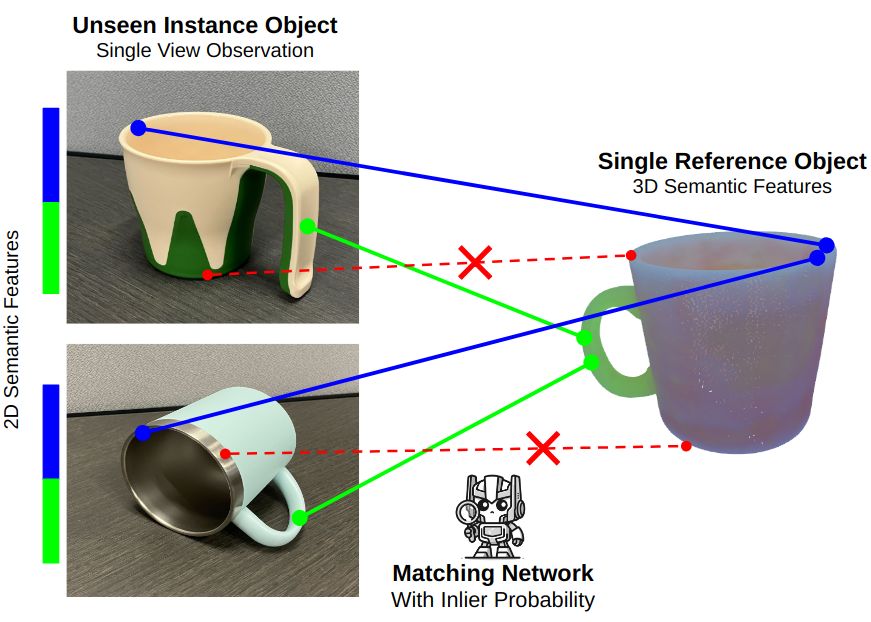}
\caption{We propose a novel approach to category-level pose estimation that makes use of 3D semantic features from a pre-trained foundation model. For a single reference object per category, 2D semantic features are projected into 3D space. We then train a transformer matching network which is used to estimate the pose of unseen objects in the category from a partial observation. Our approach is robust to the visual appearance of object instances.}
\label{fig:}
\end{figure}

In contrast, methods based on geometry only, without making use of RGB information, have shown great performance using only synthetic data, with depth information suffering from less domain gap than RGB. This means the synthetic data needs to only cover the distribution of shape variety, not texture and color.
For instance, CPPF \cite{You2022CPPFTR} use only synthetic depth information to train their network and can estimate 9D object poses  effectively on real images at test time.
However, relying on geometric information alone is not adequate to solve all ambiguities present in the pose estimation problem. For example, observing semantically meaningful parts of an object, such as the keyboard or display of a laptop, should help disambiguate the pose, even if the difference in geometry is minimal.

To mitigate the issues caused by using raw RGB or only geometric information, we propose to use semantic features provided by a pre-trained foundation model. DINOv2 \cite{oquab2023dinov2}, a self-supervised vision transformer (ViT), is one such model that can extract meaningful semantic features from RGB images. Being a transformer architecture, the features are able to capture global information in the image. 
Furthermore, the self-supervised training method enables the features to capture an understanding of object parts that are robust to texture and appearance changes. We propose to project the 2D semantic features from a category-level reference object into 3D space by sampling a few camera poses around the object and fusing the predicted features in 3D space by averaging them directly. From this, we obtain 3D semantic features for one specific object category. 

To estimate the pose of an unseen instance of this category from a single view observation, we employ a feature matching based approach. While a single-view image contains only partial information about the the target object, we hypothesize that if we have full 3D features available for the category reference object, it will be possible to find a reasonable correspondence between the partial features of the target and the full 3D features of the reference object. To aid with this, we propose a transformer matching network with inlier probability predictions, greatly assisting with matching between partial and full features. We demonstrate that our approach performs better than matching directly with raw features.

Our proposed method uses only synthetic data without requiring a large variety of 3D models for training and maintains performance when tested on real scenes thanks to the robustness of the 3D semantic features. This is in direct contrast to prior approaches that often require real data with pose annotations for training. The contributions of this paper are as follows:
\begin{enumerate}
    \item \label{contrib:dino} A novel geometric and semantic representation is introduced which greatly improves the performance for category-level pose estimation.
    \item \label{contrib:vit} We propose a robust transformer matching network for dense correspondences between partial and full information in 3D space.
    \item \label{contrib:eval} We conducted rich evaluations and compared ours with various methods. 
    These results show ours is a simple yet effective approach to category-level pose estimation, with high performance and data efficiency. 
\end{enumerate} 

\section{Related Work}

\subsection{Category-Level Object Pose Estimation}
 In the past few years, instance-level object pose estimation based on Deep Neural Networks (DNNs) has made great progress in computer vision and robotics fields \cite{Peng2019-wx, He2022-wy, He2023-tw, Liu2022-jp, pan2023learning, Nguyen2023-lb, Zhao2022-ej, Wang2020-an, Wang2021-oo, Wang_2021_self6dpp, Iwase2021-fk, Xu2022-ts, Chen2023-cu, Hodan2020-wy, Labbe2020-xp, Ikeda2022-qs}. 
On the other hand, several methods of category-level object pose estimation have recently been proposed to handle unseen object instances in specific categories without re-training \cite{Wang2019-hy, Lin2020-ai, Chen2021-qw, Lin2021-gh, Di2022-wb, Zheng2023-kn, Wen2022-fm, You2022CPPFTR}. 
There are two primary approaches to category-level 6D pose estimation. 
One approach is to estimate object pose from only observed information, such as RGB-D or depth images from a camera at test time. 
NOCS \cite{Wang2019-hy} uses only RGB as input for a Fully Convolutional Network (FCN) that is trained to estimate a Normal Object Coordinate Space (NOCS) image. Depth information is then used to estimate pose and size. 
Several methods \cite{Lin2020-ai, Chen2021-qw, Lin2021-gh, Di2022-wb, Zheng2023-kn} improve the NOCS idea based on new post-process and architecture with RGB-D inputs.
FS-Net \cite{Chen2021-qw} and GPV-Pose \cite{Di2022-wb} predict object poses directly without post-processing from RGB-D or only depth as input.
This approach can be applied to a wider range of situations since only RGB-D or depth images are required at test time.
However, such methods could stand to benefit from incorporating available information priors.
As such, the second common approach is to estimate object pose with some prior information, such as a shape prior, in addition to RGB-D at test time. 
Tian \textit{et al.} \cite{Tian2020-fd}, SPD \cite{Wang2021-jr}, RePoNet \cite{Fu2022-cz} leverage a category shape prior 
to improve performance since the shape prior contains the object's full shape information, which is difficult to estimate from a single observation.
Our approach builds on this direction, since we use a single 3D CAD model per category as a reference.

\subsection{Representation for Object Pose Estimation}
The type of input and representations used in object pose estimation is still an open research topic. For example, NOCS \cite{Wang2019-hy} uses only raw RGB images as input for the network. 
This raw RGB-based approach tends to require real data with pose annotation or a photorealistic synthetic dataset with tuned parameters for training since the dataset needs to cover the large variety of instance appearances that household objects might have within the same category. 
To further improve performance, several methods \cite{Chen2021-qw, Lin2021-gh, Di2022-wb, Zheng2023-kn} add depth information in addition to raw RGB as inputs. 
While this can improve the performance, covering both the appearance and shape variety in the training data is still essential to produce a robust model.
To solve this problem, CPPF \cite{You2022CPPFTR} uses only depth information for category-level pose estimation since the depth space is less varied compared with color space within a category \cite{Wang2021-jr}. 
This method was effective using only synthetic datasets, combined with a novel sim2real transfer technique. However this approach still requires a wide variety of object shapes in the training data, which our approach aims to reduce by incorporating semantic information.
Our approach uses both geometric and semantic information, without raw RGB, as input for our transformer matching network. Unlike prior methods, our approach incorporates 2D semantic features estimated from RGB using DINOv2 \cite{oquab2023dinov2}, projected to 3D space. 
Thanks to geometric and semantic features, only synthetic data created with a small number of instances in a category are required compared with CPPF \cite{You2022CPPFTR}.

To the best of our knowledge, our approach is the first to use semantic features in 3D space to achieve category-level pose estimation with only a small number of training objects.

\section{Method}

\begin{figure*}[!ht]
\centering
\includegraphics[width=1.0\linewidth]{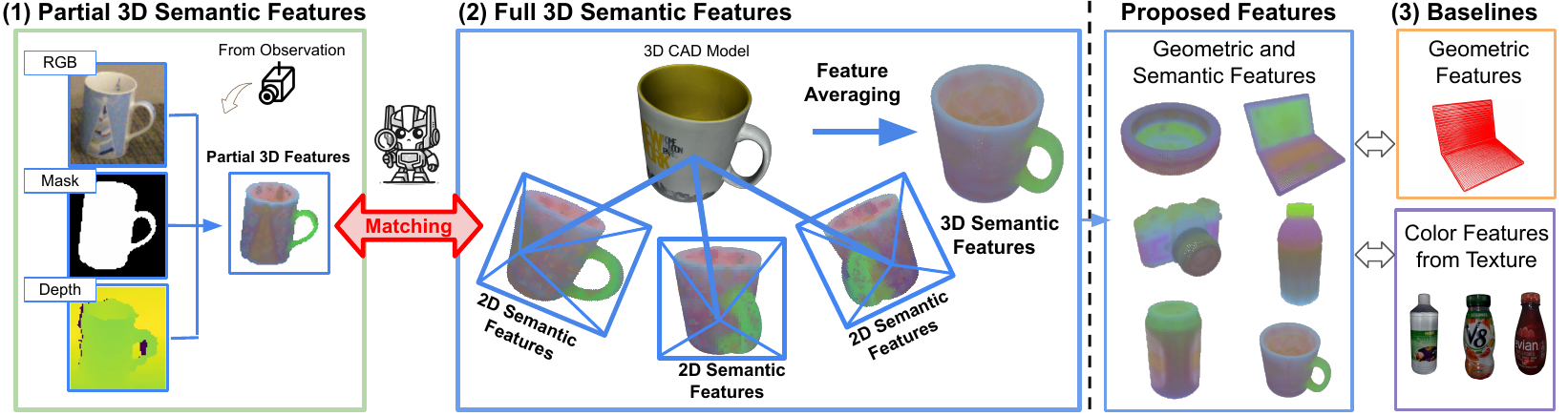}
\caption{ \textbf{Overview of our pipeline}. Different from other category-level pose estimation pipelines, our method incorporates both geometric and semantic features to improve performance. Specifically, we fuse 2D semantic features from multiple sampled views of synthetic object models to generate 3D semantic features for each category. During the inference stage, RGB images and object masks are used as inputs to obtain the partial point cloud with semantic features. The 9D object pose is retrieved by performing dense matching between (1) the partial 3D observation to (2) the full 3D semantic features. In comparison, (3) baseline methods such as CPPF \cite{You2022CPPFTR} only utilize geometric features, while others (NOCS \cite{Wang2019-hy}) leverage RGB images and need a large amount of textured objects for the training. In contrast, our network requires much fewer training objects with a good performance with the novel semantic representation in 3D space.}
\label{fig:teaser1}
\end{figure*}

We assume a set of synthetic CAD models $S= \{ S_i, | i = 1,\cdot \cdot \cdot,N \} $ are available in one category during training. 
Given a RGB-D image with detection mask of a novel instance for this category at inference time, our task is to recover the 9D object pose including the rotation $R$, translation $t$ and scale $s$, assuming access to a single reference CAD model from the set $S$.

\begin{figure*}[t]
 \centering
    \includegraphics[width=1.0\linewidth]{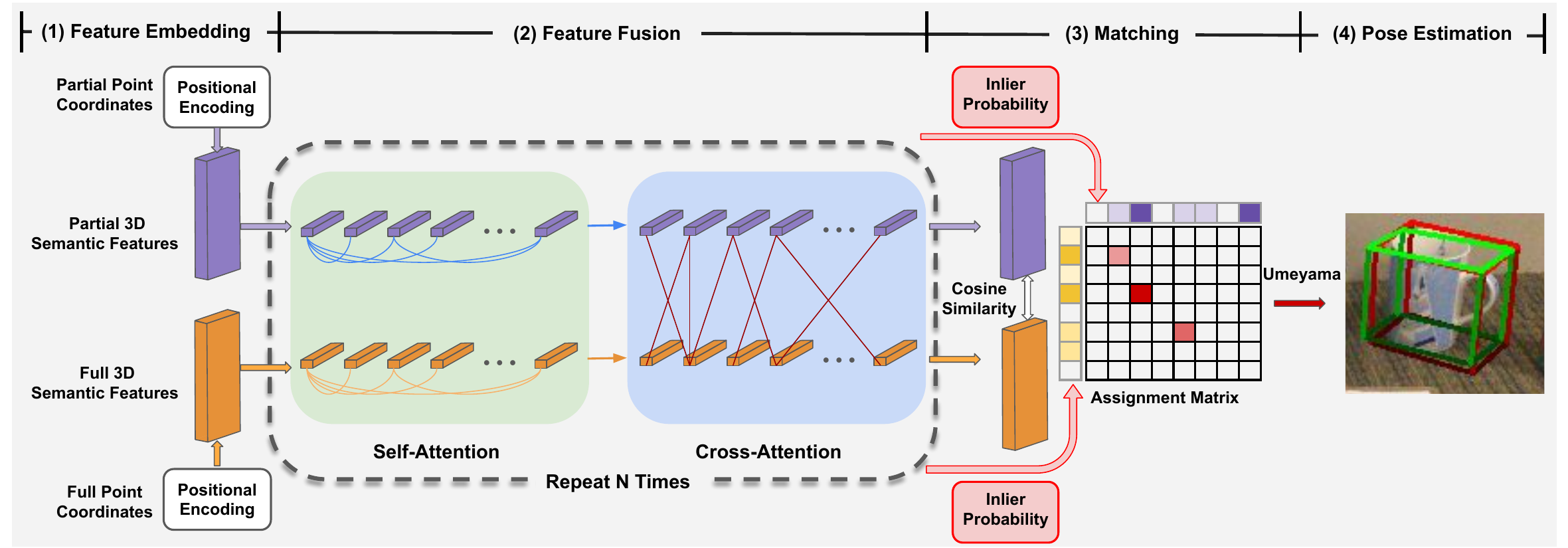}
    \caption{ \textbf{Overview of our transformer matching network}. To match partial input and full model points with semantic features, (1) we first embed normalized point coordinates as geometric features with positional encoding and add them with semantic features. (2) The embedded features are fused with self- and cross-attention layers for multiple iterations for global perceptions. (3) We predict the inlier probability for both partial 3D semantic features and full features, and multiply them in the assignment matrix from cosine similarities to reduce outliers. (4) Finally, 9D object poses of novel instances are retrieved by Umeyama algorithm \cite{Umeyama1991LeastSquaresEO} with RANSAC \cite{Fischler1981RandomSC} from the dense correspondences.}
    \label{fig:teaser2}
\end{figure*}

Different from previous methods, no real images or pose annotations are available during training. Further, the number of synthetic objects in our dataset are greatly reduced when compared to that of prior works. %
The overall pipeline of our proposed method is visualized in Fig. \ref{fig:teaser1}.
Our method improves over prior synthetic-only
approaches such as CPPF \cite{You2022CPPFTR} by incorporating semantic information obtained from a 2D image foundation model, in the form of features which are agnostic to  specific visual appearances of objects within a category. 
We leverage these features to create 3D semantic features from multiple rendered views of the reference object by projecting the image features into the object point cloud.
As shown in Fig. \ref{fig:teaser2}, a transformer is utilized to fuse both partial 3D semantic features and full features 
for accurate 3D matching during inference. This transformer is trained on only a small number of CAD models for the given category and is able to recover dense correspondences, allowing us to recover the 9D object pose.

\subsection{Recap on Feature Embeddings}

Semantic information is extremely beneficial for disambiguating object poses. Instance-level methods such as DenseFusion \cite{wang2019densefusion} leverage fused RGB and depth images jointly and improve the pose estimation performance. However, object textures can vary for different instances within a category. As a result, it is challenging to generalize to new textures on novel instances in the real scenes. Therefore, some methods \cite{Chen2021-qw, Di2022-wb, Zhang2022RBPPoseRB}
directly exploit the geometric similarity from depth information. Similarly, the CPPF \cite{You2022CPPFTR}, which is the SOTA synthetic-only approach, adapts point pair features (PPF) encoding local patch information to guide the pose as well as scale predictions. While being effective on certain categories such as bottles or cans, the method struggles with categories such as mugs or laptops. The challenge is ambiguous local geometry structures, for example the keyboard and screen of a laptop are geometrically similar, and mugs often have similar local geometries except for a parts of the handle which tend to be ignored due to unbalanced data. We argue that utilizing a semantic representation from a pre-trained foundation model would reduce the sensitivity to texture differences while providing vital semantic clues to help tackle ambiguous geometry structures.

Image foundation models, typically pre-trained on extensive and diverse datasets, provide a powerful base model, the features from which are often powerful at extracting semantic features. 
With the prevalence of the transformer architecture, foundation models based on vision transformers such as DINOv2 \cite{oquab2023dinov2} are able to better capture global relationships inside the features.
However, using these features to perform matching from 2D image pairs has limitations. Feature registration from 2D partial observations works when the observations are from a similar viewing angle. Additionally, it is challenging to cover full 360\degree \! views of the object in the inference stage. 
In contrast, matching features from a partial view to full 3D features is a promising direction. However, 3D foundation models 
are still yet to be thoroughly explored, because of the challenge in collecting large scale 3D assets for training. To tackle this challenge, we reuse the 2D foundation models and project the features to the object point cloud to be used for 3D-3D matching.

\subsection{Semantic Feature Wrapping}

As shown in Fig. \ref{fig:teaser1}, to lift the 2D semantic features to 3D, we first sample camera poses around the objects, ensuring the model points are visible in at least one view. Next, the rendered RGB images are transformed to semantic features with DINOv2 which are resized to the original image size. For each frame, the visibilities of the object vertices in the mesh are calculated. Based on the camera pose and intrinsics, the visible points are projected to the 2D feature image to retrieve the corresponding semantic features. To align the feature discrepancies from multiple observations, we take the average of the visible features from multiple views for each point. The averaging additionally filters the noise from multiple predictions, shown in (2) from Fig. \ref{fig:teaser1}.

The properties of 2D semantic features are preserved after lifting into 3D space. For example, the features can be clustered for zero-shot object part semantic segmentation tasks. 
However it is challenging to do matching directly from the partial 3D semantic features to the full object reference 3D features for pose estimation since it requires high-quality correspondence matching.
Despite robustly encoding salient object semantics such as mug handles, the semantic features have minor differences inside semantically similar regions which leads to noisy matching results. Visual discrepancies of different instances in one category also result in matching outliers. The above matching outliers are hard to filter manually and lead to inferior pose estimation results. Therefore, we propose to fuse the semantic and geometric features jointly as input to a transformer matching network for accurate pose estimations.

\subsection{Transformer Matching Network}
Given partial 3D semantic features from observation and full features from one reference CAD model, the task is to find accurate correspondences between them. Inspired by OnePose \cite{Sun2022-my, He2023-tw}, we utilize a transformer structure with multiple self- and cross-attention layers to fuse both semantic and geometric features, as shown under (2) in Fig. \ref{fig:teaser2}. Specifically, the geometry features are embedded with the positional encodings of point coordinates and concatenated with the semantic features as network inputs, visualized under (1) in Fig. \ref{fig:teaser2}. We assume the partial inputs $P$ and full inputs $Q$ have $M$ and $N$ points, with corresponding features $F^P$ and the features $F^Q$ before fusion. As shown in Fig. \ref{fig:teaser2}, the features are fused for both partial 3D semantic features and full features inside the self-attention layers. 
Followed by the cross-attention layer between the partial and full features, the global features across both inputs can be learned in addition to the local features. After fusion the features are denoted as $\hat{F}^P$ and fusion $\hat{F}^Q$.

Matching between features from partial observations and the full 3D reference features has an advantage over simply matching between only partial observations in that the full features are available for reference.
However, this may additionally increase the potential for mismatches to occur as the possible matching regions are also expanded.
To avoid matching to regions that are out of input view visibility, we find it empirically useful to predict the inlier probability from the output features similar to LightGlue  \cite{Lindenberger2023-eo}. The prediction of inlier probabilities helps the network to constrain the target matching regions and guides the feature representation learning inside the attention regions.

To summarize, the assignment matrix A of shape $M \times N$ is calculated by the cosine similarity between $\hat{F}^P$ and $\hat{F}^Q$. The inlier probabilities are predicted as $\sigma^P$ and $\sigma^Q$ for the partial and full features.  Afterwards the assignment matrix $\hat{A}$ by the inlier probabilities is obtained by multiplication of the cosine similarity in Equ. \ref{equ:0}.
{\setlength{\mathindent}{0cm}
\begin{align}
     \hat{A}_{i,j} & = \sigma^P_i  \cdot \sigma^Q_j \cdot A_{i,j}  \quad \forall(i,j) \in P \times Q  \label{equ:0} \\
    L_{P} = - \frac{1}{|P|} & \sum_{i \in P}  ( \sigma_{i,gt}^P log\sigma_i^P +(1-\sigma_{i,gt}^P) log(1-\sigma_i^P) ) \label{equ:1}  \\
 L_{Q} = - \frac{1}{|Q|} & \sum_{j \in Q}  (  \sigma_{j,gt}^Q log\sigma_j^Q +(1-\sigma_{j,gt}^Q) log(1-\sigma_j^Q)) \label{equ:2} 
\end{align}

The training loss is the sum of inlier classification losses and the assignment matrix loss. The inlier classification losses are in Equ. \ref{equ:1} for partial inputs $P$ and Equ. \ref{equ:2} for full inputs $Q$. The assignment matrix loss is calculated in Equ. \ref{equ:focal} with the focal loss \cite{lin2017focal} and $\gamma$ as 2. $A_{pos}$ and $A_{neg}$ are the positive and negative ground truths for the assignment matrix $A$.
{\setlength{\mathindent}{0.9cm}
\begin{equation}
\begin{split}
    L =  &  -\frac{1}{ |A_{pos}| } \sum_{\hat{A}_{i,j} \in A_{pos}} (1-\hat{A}_{i,j})^\gamma \log (\hat{A}_{i,j}) \\ 
    & - \frac{1}{ |A_{neg}| } \sum_{\hat{A}_{i,j} \in  A_{neg} } \hat{A}_{i,j}^\gamma \log (1-\hat{A}_{i,j})
\end{split}
\label{equ:focal}
\end{equation}}

Based on the similarity matrix from the output features, a threshold is applied to extract high confidence matches. To recover the 9D pose of novel instances, it is assumed that object shapes in the same category share the same topology. Therefore, most of the novel instance shapes can be approximated via affine transformations from the shape prior. To this end, Umeyama algorithm \cite{Umeyama1991LeastSquaresEO} combined with RANSAC algorithm \cite{Fischler1981RandomSC} is applied based on the accurate matched pairs to robustly recover the rotation, translation and the object scales.

\subsection{Disambiguating Symmetry}

For many objects, there exist symmetries that cause ambiguities in the object pose, where the network will be trained against conflicting ground truth signals for a given pose. This presents a significant challenge in the pose estimation problem.

Therefore, we extract unique ground truth poses by constraining the object xz-plane, (red and blue-axis plane as shown in the Fig. \ref{fig:symmetry}) to always intersect with the origin of camera coordinate system.
We also treat the mug as an axis-symmetry object when the handle is invisible in the view.

\begin{figure}
\centering
\includegraphics[width=1.0\linewidth]{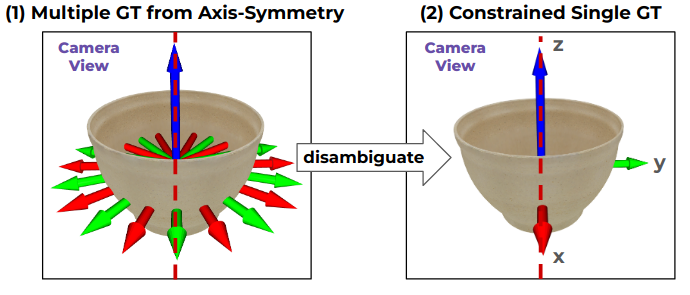}
\caption{\textbf{Disambiguating the symmetrical poses}. %
(1) Since multiple ground truth poses can exist for axis-symmetry objects, (2) the Ground Truth(GT) pose is constrained to intersect the object xz-plane with the camera origin coordinate system.}
\label{fig:symmetry}
\end{figure}

\section{Experiments} \label{exp}

To cover as many categories and novel instances 
as possible, three datasets: NOCS\cite{Wang2019-hy}, Wild6D \cite{Fu2022-cz} and SUN RGB-D \cite{Song2015SUNRA} are chosen for the testing. NOCS covers 18 instances from 6 categories and Wild6D includes 162 objects from 5 categories. SUN RGB-D contains 10182 9D bounding boxes for the chair category in indoor environments. All the datasets provide RGB-D images and the corresponding 9D pose annotations.

\subsection{Implementation Details}
To evaluate the performance of our method against other synthetic-only approaches, we directly train our model on synthetic objects from the ShapeNet dataset \cite{Chang2015ShapeNetAI} and test on the three real datasets \cite{Wang2019-hy, Fu2022-cz, Song2015SUNRA} that are mentioned in section \ref{exp}.
Only 10 ShapeNet models are selected from corresponding categories.
For each object, 40 images from different views are rendered for the generation of 3D semantic features and used as the synthetic training dataset for the transformer matching network. The network is trained with a learning rate of 1e-4 for 100 epoches for each category on a desktop with Intel Xeon E5-2698 CPU and Tesla V100-DGCS-32GB GPU. The networks trained on bottle, bowl, camera, can, laptop and mug categories are used for the evaluation on the NOCS and Wild6D datasets. The network trained on chairs is used for the evaluation on the SUN RGB-D dataset.

\subsection{Metrics}

\begin{figure*}[t]
 \centering
    \includegraphics[width=\linewidth]{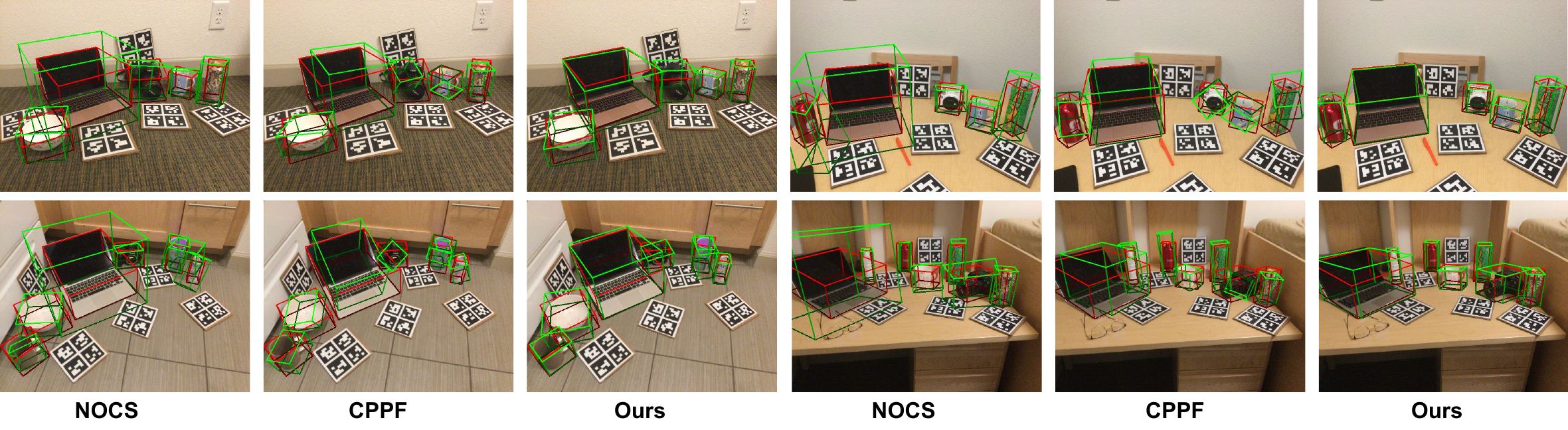}
    \caption{Visualizations of predicted 3D bounding boxes from NOCS \cite{Wang2019-hy} (left), CPPF  \cite{You2022CPPFTR} (middle), and ours (right) for scenes in the NOCS REAL275 dataset. Green is predicted and red is the ground truth.}
    \label{fig:}
\end{figure*}

\begin{figure}[t]
\centering
\begin{tabular}{cc}
\includegraphics[width=0.45\linewidth]{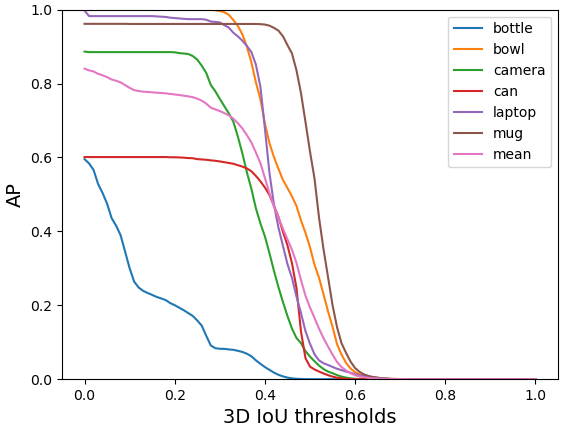}&
\includegraphics[width=0.45\linewidth]{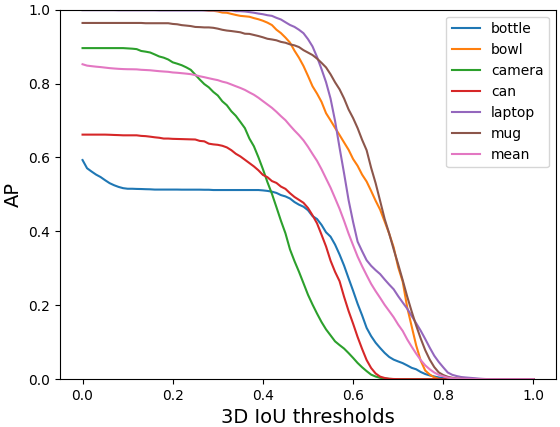}\\
\end{tabular}
\caption{Visualization of each category's 3D IoUs for CPPF \cite{You2022CPPFTR} and Ours on the NOCS REAL275 dataset.} 
\label{fig:vis_iou}
\end{figure}

For the evaluation on the NOCS and Wild6D dataset, the mean precision of 3D intersection over union (IoU) at thresholds of 25\%, 50\% \cite{You2022CPPFTR, Fu2022-cz} are reported for jointly evaluating rotation, translation and size. The $5 \degree 5$cm, $10 \degree 5$cm, $15 \degree 5$cm metrics are leveraged to measure the accuracy of rotations and translations \cite{Wang2019-hy}. For the evaluation on the SUN RGB-D dataset, the mean precision of 3D intersection over union (IoU) at thresholds of 10 \%, 25\% are used. The $20 \degree 10$cm, $40 \degree 20$cm, $60 \degree 30$cm metrics are used for the evaluation of rotation and translation error.

\subsection{Performance Analysis}

\textbf{Performance on  NOCS REAL275 Dataset} The NOCS REAL275 dataset collects the object pose annotations of six categories, with 8K images among 18 real scenes in total. We utilize the testing split including 2750 images for the evaluation. The evaluation result in comparison with SOTA baselines with synthetic-and-real and synthetic-only approaches
are listed in Tab. \ref{tab:test_nocs}. 
In comparison with DualPoseNet \cite{Lin2021-gh} 
trained with the real data, our method shows comparable results for the 3D IoU, rotation and translation metrics by using only synthetic data for training. 
The $3D_{25}$ and  $15 \degree 5$cm scores are close, while ours outperforms DualPoseNet on $3D_{50}$ by 5.9\%. The $5 \degree 5$cm and $10 \degree 5$cm scores are slightly lower than the baseline, which are caused by small canonical coordinate differences defined in ShapeNet objects and shape modelling discrepancies through affine mesh transformations. 
In comparison with synthetic-only approaches, our method leads to an overall increase of 3D IoU, rotation and translation scores. Especially the $3D_{50}$ metric increases greatly by 36.8\%, even though the method is trained on a smaller amount of synthetic objects. The $5 \degree 5$cm, $10 \degree 5$cm, $15 \degree 5$cm increase by 11.9\%, 15.2\%, 22.8\%. It is observed that ours performs better than CPPF on difficult categories such as mugs and laptops.
Our method takes 3D semantic features encoding object semantics and fuses the global information in a transformer architecture, which boosts its performance on challenging categories. Another example is that CPPF needs to train additional classifier for laptop category to determine the top and bottom part because of local geometry ambiguity, which is unnecessary for our method.    

\mytableA

\begin{figure}[t]
\centering
\begin{tabular}{cc}
\includegraphics[width=0.45\linewidth]{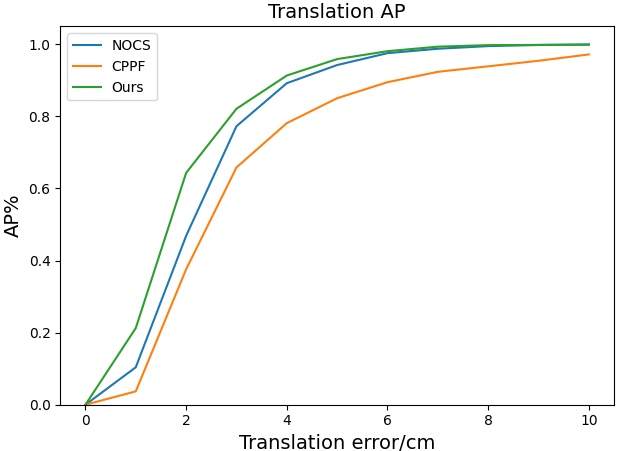}&
\includegraphics[width=0.45\linewidth]{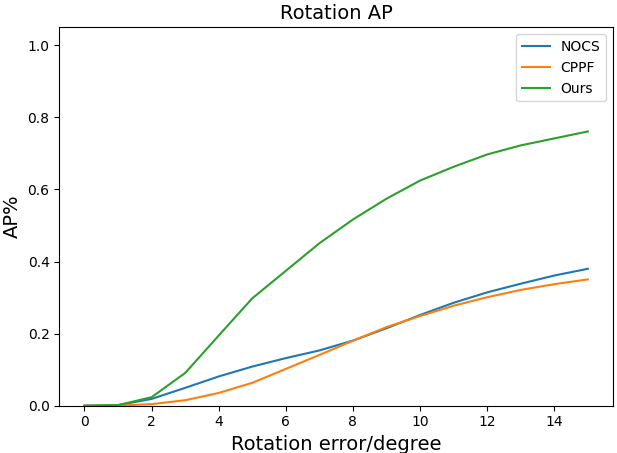}\\
\end{tabular}
\caption{Visualization of translation and rotation mAPs for our method in comparison with CPPF\cite{You2022CPPFTR} and NOCS \cite{Wang2019-hy} on the NOCS REAL275 dataset.} 
\label{fig:vis_map}
\end{figure}

\mytableD

\mytableE

\textbf{Performance on Wild6D Dataset} The Wild6D datasets contains 5166 videos over 1722 object instances among 5 categories, of which 486 videos over 162 objects are leveraged for the testing. The number of testing object instances are a magnitude higher than the NOCS REAL275 dataset, which better reflects the texture and shape distributions of category-level objects in the real world, and poses a challenge to the model generalisation ability.  
The evaluation results are shown
in Tab. \ref{tab:test_wild6d}. In comparison with methods trained with real data such as DualPoseNet, our method provides comparable results for the  $3D_{25}$ (84.6\% vs 90.0\%), $3D_{50}$ (67.7\% vs 70.3\%), $5 \degree 5$cm (30.8\% vs 34.4\%) metrics, and outperforms the state-of-the-art method on  $10 \degree 5$cm by 8.5 \% without using real data. 
The result shows that even trained on a small amount of synthetic object models, our method generalizes to a large variety of object shapes and textures in the real scenes. Detections are also visualized in Fig. \ref{fig:vis_wild}.
Failure cases are observed when the depth estimations fail for transparent bottles, or inaccurate 2D segmentations of the cameras lead to the pose estimation failures. 

\begin{figure}[t]
\centering
\begin{tabular}{cc}
\includegraphics[width=0.45\linewidth]{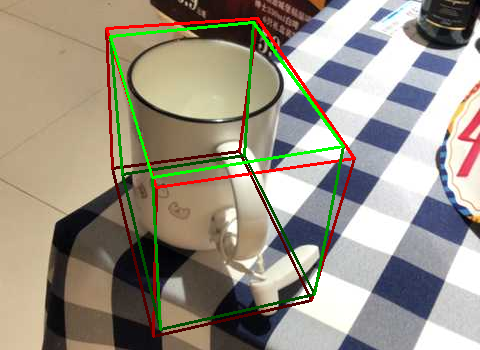}&
\includegraphics[width=0.45\linewidth]{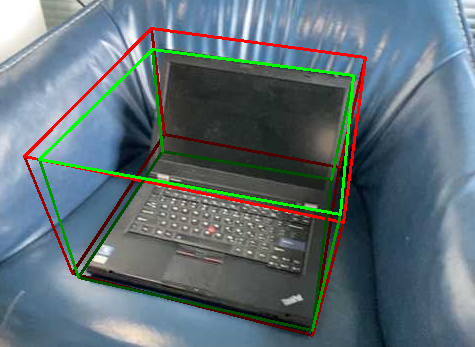}\\
\end{tabular}
\caption{Visualization of predicted 3D bounding boxes on Wild6D dataset \cite{Fu2022-cz}. Green is predicted, red is the ground truth.} 
\label{fig:vis_wild}
\end{figure}

\textbf{Performance on SUN RGB-D Dataset}
Beyond a variety of household objects contained in NOCS and Wild6D dataset, we further evaluate our model on challenging indoor scenes in SUN RGB-D dataset, where there are huge object shape discrepancies of the chair category in ShapeNet and SUN RGB-D dataset. We evaluate on all the chairs in the validation dataset following the setup in CPPF \cite{You2022CPPFTR}. 
The evaluation results are shown in Tab. \ref{tab:test_sunrgbd}, and our proposed method outperforms the baseline by 20.8\% and 18.4\% on $3D_{10}$ and $3D_{25}$ metric, which shows good generalization ability towards zero-shot object poses estimation in indoor scenes. The rotation and translation scores are higher than the baseline, especially for $40 \degree 20$cm and $60 \degree 30$ cm. The $3D_{25}$ metric is lower than the categories in the Wild6D dataset \cite{Fu2022-cz} because of the heavy occlusions in the indoor scenes, for example the chairs are hidden by the tables or only partial visible in the image corner. It is observed that our method is still robust under partial occlusion, as shown in Fig. \ref{fig:vis_sun}.

\begin{figure}[t]
\centering
\begin{tabular}{cc}
\includegraphics[width=0.45\linewidth]{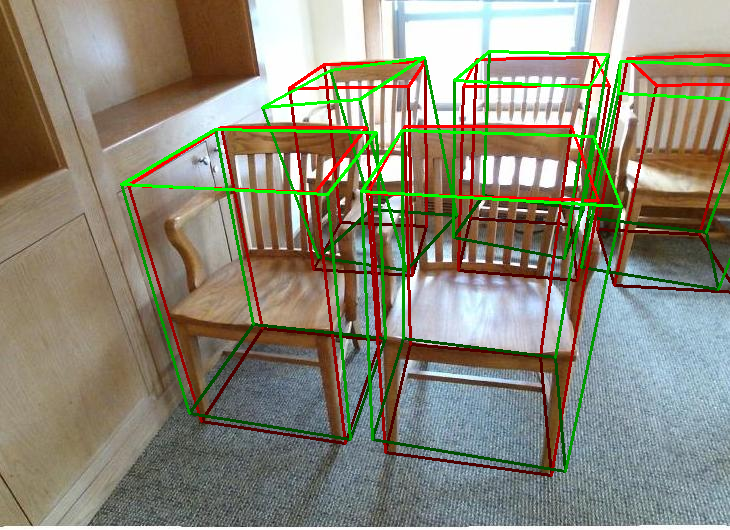}&
\includegraphics[width=0.45\linewidth]{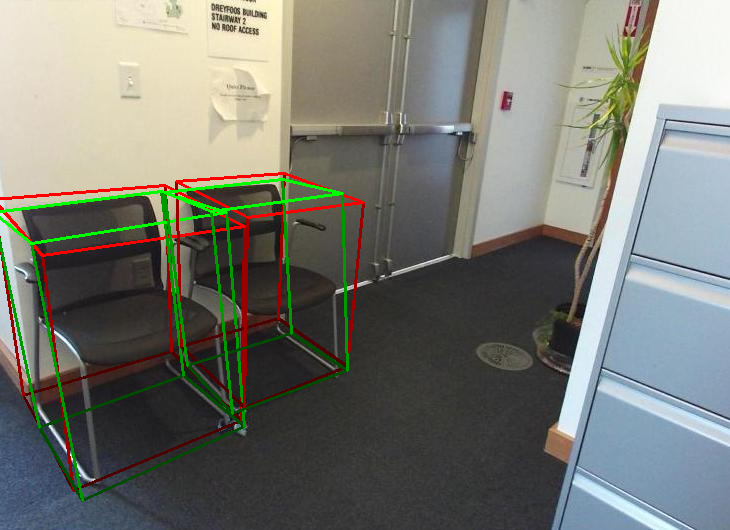}\\
\end{tabular}
\caption{Visualization of predicted 3D bounding boxes on SUN RGB-D dataset \cite{Song2015SUNRA}. Green is predicted, red is the ground truth.} 
\label{fig:vis_sun}
\end{figure}

\subsection{Ablation Study}

To analyse the influence of different network components and settings, exhaustive ablations are performed and the following results are reported for the NOCS dataset.

\mytableC

\mytableB 

\begin{figure}[t]
\centering
\includegraphics[width=0.95\linewidth]{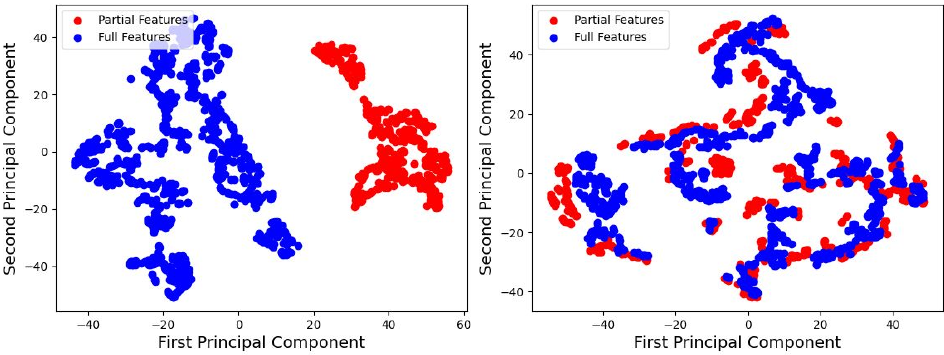}
\caption{t-SNE visualization of 3D semantic features from partial 3D features (red) and full 3D features (blue)
inside the attention region before and after feature fusion.}
\label{fig:vis_tsne}
\end{figure}

\textbf{Influence of Feature Fusion Network}
We hypothesize that performing 3D-3D matching with these features directly without the feature fusion network, labelled as (2) in Fig. \ref{fig:teaser2}, performs poorly since the 3D semantic features provided by DINOv2 lack precise local information.
We demonstrate this by ablating our matching network, and instead directly performing matching between the partial 3D semantic features and the full reference features with the maximum cosine similarity. 
Results of this ablation study are shown in Tab. \ref{tab:test_ablation}. 
The $5 \degree 5$cm, $10 \degree 5$cm, $15 \degree 5$cm scores shows inferior results and the $3D_{25}$, $3D_{50}$ are low with values of 2.2\%, 0.1\% respectively, due to outliers and inaccurate matches causing scale estimation error in the Umeyama algorithm.
Unlike the matching of semantic features between 2D images, matching between partial 3D features and full 3D features is more challenging as there are more potential semantic match candidates.
The result shows the necessity of our fusion step to refine the matches between the partial 3D semantic features and full features with the strengths of both geometric and semantic features for 9D pose estimations.
As shown with the t-SNE visualization example from Fig. \ref{fig:vis_tsne}, the partial 3D semantic features and full features are separately distributed before the fusion. Despite this, the features of both sides are well aligned after the fusion process, which explains the failures of directly matching the raw features.

\textbf{Inlier Probability Prediction for Matching}
In the ablation  $A_2$ in Tab. \ref{tab:test_ablation}, the inlier probability module is removed, including calculation of the assignment matrix (Equ. \ref{equ:0}) and the inlier classification loss (Equ. \ref{equ:1} and \ref{equ:2}). 
Without consideration of matching inliers, the $3D_{25}$, $3D_{50}$ drop by 17.2\% and 14.3\%. In addition to the worse 3D bounding box predictions, the rotation and translation scores also decrease slightly.  The $5 \degree 5$cm, $10 \degree 5$cm, $15 \degree 5$cm decrease by 8.9\%, 10.8\%, 6.7\%. Evaluation shows that the inlier probability prediction module is crucial in the partial to full feature matching process. The mechanism helps the network to focus on the regions of attention and reduces the outliers in the final matching stage.

 \textbf{Symmetric Handling}
 In the ablation $A_3$, we remove the symmetry handling of all categories in the training. The result shows that the $3D_{25}$ drops slight to 72.3\%, while $3D_{50}$ decreases greatly to 44.5\%. The $5 \degree 5$cm, $10 \degree 5$cm, $15 \degree 5$cm drop by 16.1\%,  19.6\%, 16.9\%. The  results show that the conflicting ground truth matches confuse the network and lead to inferior performances, and it is important to disambiguate the ground truth matches for the axis-symmetry categories. 

\textbf{Influence of Synthetic Training Object Numbers}
To show the influence of the number of synthetic objects for training, we train CPPF with different object numbers and show the evaluation result in Tab. \ref{tab:test_objnum}. 
The $5 \degree 5$cm, $10 \degree 5$cm, $15 \degree 5$cm score increases with the number of training objects, which shows that approaches such as CPPF that rely on geometric information and synthetic-only data require more object shape variation in the training dataset for better generalization capability.

In contrast, we train our method with 10, 20, 40 synthetic models as shown in Tab. \ref{tab:test_objnum}. The result shows that the performance saturates already with as few as 10 objects on $3D_{25}$ and $3D_{50}$. In comparison to depth inputs, fusing semantic information from 2D foundation models with geometric encodings prove to have a stronger generalization ability.
\section{Conclusion}

In this paper, we introduce a novel representation incorporating both semantic and geometric features for the category-level pose estimation task. Based on the novel representation, a transformer matching network is trained which predicts inlier probabilities and reduces matching outliers between partial and full 3D semantic features. While requiring significantly less object instances, our method outperforms baselines by a great margin and shows an outstanding generalization ability on multiple evaluation datasets. An interesting potential avenue for improvement is to additionally estimate the target object's  deformation against the reference 3D model, which could potentially improve both 3D size and 6D pose estimation. Furthermore, it is straightforward to extend our approach from single observations to multiple observations. We leave these directions for future work.

{\small
\bibliographystyle{ieeenat_fullname}
\bibliography{egbib}
}

\end{document}